# An Effect of Spatial Filtering in Visualization of Coronary Arteries Imaging.

**Dr. P.S. Hiremath[1], Mr. Kodge B.G.[2]**
1. Professor & Chairman, Department of Computer Sci. Gulbarga University, Gulbarga.
State: Karnataka (INDIA)
hiremathps@hotmail.com
2. Lecturer, Department of Computer Science, S. V. College, Udgir - 413517, Dist. Latur
State: Maharashtra (INDIA)
kodgebg@hotmail.com

## ABSTRACT

*At present, coronary angiography is the well known standard for the diagnosis of coronary artery disease. Conventional coronary angiography is an invasive procedure with a small, yet inherent risk of myocardial infarction, stroke, potential arrhythmias, and death. Other noninvasive diagnostic tools, such as electrocardiography, echocardiography, and nuclear imaging are now widely available but are limited by their inability to directly visualize and quantify coronary artery stenoses and predict the stability of plaques. Coronary magnetic resonance angiography (MRA) is a technique that allows visualization of the coronary arteries by noninvasive means; however, it has not yet reached a stage where it can be used in routine clinical practice. Although coronary MRA is a potentially useful diagnostic tool, it has limitations. Further research should focus on improving the diagnostic resolution and accuracy of coronary MRA. This paper will helps to cardiologists to take the clear look of spatial filtered imaging of coronary arteries.*

### *KEY WORDS*
Visualization, Spatial Filtering, Coronary vessels, Angiography, MRA.

## 1. INTRODUCTION

The term spatial domain means the image plane (Flat) itself, and the different methods in this category are based on direct manipulation of pixels in coronary arteries images.[6] Here we are going to focus attention on image inversing (negative), Laplacian filtering & unsharp masking, spatial convolution, Edge detection and image arithmetic, and some other methods are applied on magnetic resonance images of coronary artery.

The visualization of Magnetic Resonance imageries of coronary arteries with applying the different kinds of spatial filtering techniques will increase the accuracy and clearness to identify the cardiac diseases. Coronary magnetic resonance angiography (MRA) is a technique that allows the visualization of coronary arteries by non-invasive means.[1]

Ischemic heart disease is the leading cause of death worldwide. For patients with coronary artery disease (CAD), treatment consists of medical therapy, percutaneous intervention, or coronary artery bypass grafting. The most essential diagnostic tool to evaluate graft disease is coronary angiography. This technique provides rapid and accurate delineation of the entire coronary anatomy including the graft, and the native non-grafted coronary arteries, and it is generally used as a prelude to revascularization procedures. [2]The potential benefit of coronary MRA is not only the visualization of coronary arteries, but also the visualization of the cardiac morphology, acquired diseases of the great vessels, and cardiac function at rest and under stress[3–5]. With the recent introduction of multirow detector systems, helical computed tomography (CT) imaging of the heart and MRI has gained renewed interest[9].

## 2. Spatial Filtering

Coronary MRA is currently used much less frequently than MSCT angiography to detect CAD. Images obtained with coronary MRA provide a measure of stenosis, wall thickening and possibly composition of the plaque (Fig. 1). Coronary MRA captures images by detecting protons in moving blood every few milliseconds as they travel through an arbitrary region in space[11].

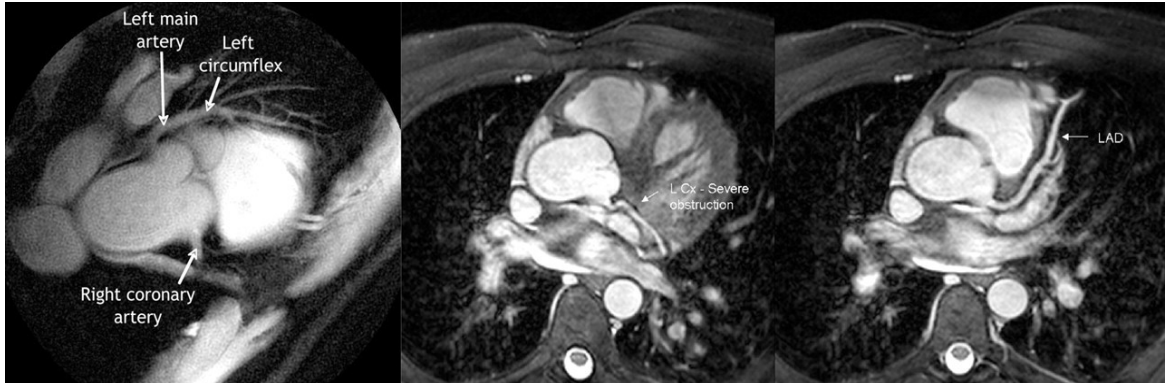

**Figure 1. Magnetic Resonance Imaging of Human Heart**

High-resolution MRI has the potential to noninvasively image the human coronary artery and define the degree and nature of coronary artery disease. Coronary artery imaging by MR has been limited by artifacts related to blood flow and motion and by low spatial resolution[8].

The spatial domain techniques are operating directly on the pixels on the image. The spatial domain processes are denoted by the expression

$$A(x, y) = T[B(x, y)] \qquad (2.1)$$

Here $B(x, y)$ is the input image, $A(x, y)$ is the output (processed) image, and $T$ is an operator on B, defined over a specified neighborhood about point $(x, y)$. In addition T can operate on a set of images. The values of pixels, before and after processing, will be denoted by r and s, respectively. The values are related by an expression of the form[6]

$$s = T(r) \qquad (2.2)$$

Where T is a transformation that maps a pixel value r into a pixel value s.

### 2.1 Image Negatives

The negative of an image with gray levels in the range $[0, L-1]$ is obtained by using the negative transformation is given by the expression

$$s = L - 1 - r. \qquad (2.3)$$

Reversing the intensity levels of an image in this manner produces the equivalent of photographic negative. Fig. 2(a) is an original digital MRI (Magnetic Resonance Imaging) of coronary artery, Fig. 2(b) is the negative image applied by an Eq.(2.3). In spite of the fact that the visual content is the same in both images Fig.2(a) and (b), note how much easier it is to analyze the coronary artery in the negative image in this particular case. And Fig. 2(c) Produces a result similar to (but with more gray tones than) Fig.2(b) by compressing the low end and expanding the high end of the gray scale.

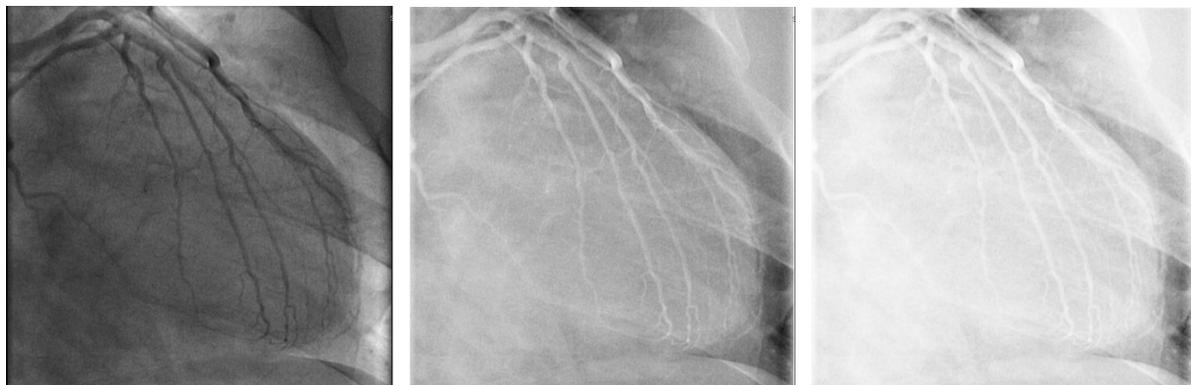

**Figure 2. (a) Original digital MRI of coronary artery. (Courtesy of Apollo DRDO hospital Hyd. INDIA) (b) Negative image obtained using the negative transformation using Eq.(2.3).**
**(c) Low end compressed and high end expanded gray scale image.**

## 2.2 Laplacian and convolution filtering

Here we are interested in isotropic filters, whose response is independent of the direction of the discontinuities in the image to which the filter is applied. It can be shown that the simplest isotropic derivative operator is the Laplacian, which for a function ( image ) $f(x, y)$ of two variables are defined as[6-7]

$$\nabla^2 f = (\delta^2 f / \delta^2 x^2) + (\delta^2 f / \delta^2 x^2) \qquad (2.4)$$

Because derivatives of any order are linear operations, the Laplacian is a linear operator. In order to be useful for digital image processing, this equation needs to be expressed in discrete form. Taking in to account that we know have two variables, we use the following notation for the digital implementation of the two-dimensional Laplacian in Eq. (2.4) is obtained by summation of x and y, direction components:

$$\nabla^2 f = [f(x+1, y) + f(x-1, y) + f(x, y+1) + f(x, y-1)] - 4f(x, y) \qquad (2.5)$$

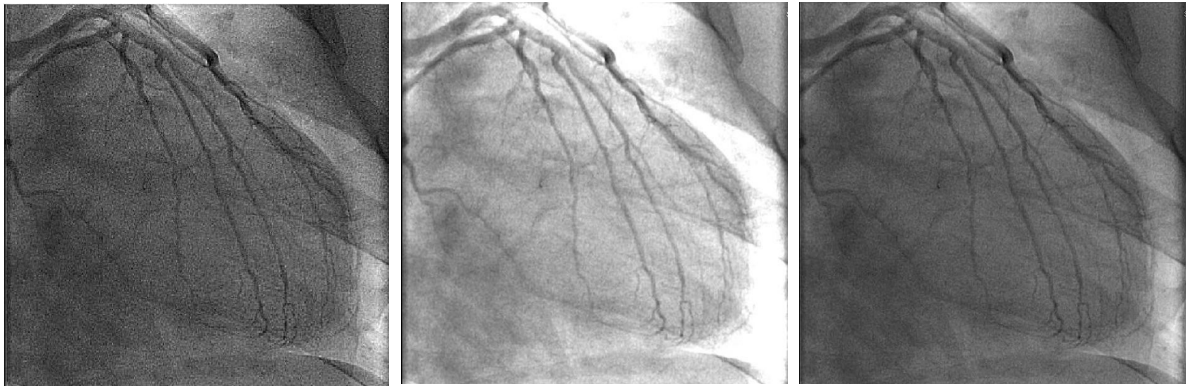

**Fig.3. (a) Laplacian filtered Image (b) Unsharp masked image (c) Image filtered by Convolution.**

Usually, the sharper enhancement is obtained by using the 3 x 3 Laplacian filter that has a -8 in the center and is surrounded by 1s, Fig. 3(a) use the Laplacian by subtracting (because the center coefficient in negative) the Laplacian image by original image. And it is for the manual implementation of this filter. The sharpen images are consists of subtracting a blurred version of an image from the image itself. This process is called unsharp masking and Fig. 3(b) is a result of unsharp masking. Unsharp masking is expressed as[7]

$$f_s(x, y) = f(x, y) - \bar{f}(x, y) \qquad (2.6)$$

where $f_s(x, y)$ denotes the sharpen image obtained by unsharp masking , ans $\bar{f}(x, y)$ is a blurred version of $f(x, y)$. The Fig. 3(c) is an output convolution filtering, and the convolution can be performed by rotating the variable w with 180º and place its rightmost point at the origin of image $f$.

## 2.3 Edge detection, image arithmetic and histogram processing.

MR angiography and CT techniques have evolved so that their technical reliability is very high for obtaining a diagnostic-quality study with adequate signal-to-noise ratio and spatial resolution and without artifacts and other limitations[10].
A problem of fundamental importance in image analysis is edge detection[12]. Edges characterize object boundaries and are there fore useful for segmentation, registration, and identification of objects in scene. Edge point can be thought of as pixel locations of abrupt gray-level change. For example, it is reasonable to define edge point in binary images as black pixels with at least one white nearest neighbor, that is pixel locations (m, n) such that u(m, n) = 0 and g(m, n) = 1, where[12]

$$g(m, n) \triangleq [u(m, n) (+) u(m \pm 1, n)]. \text{ OR } [u(m, n) (+) u(m, n \pm 1)] \qquad (2.7)$$

Where (+) denotes the exclusive-OR operation for continuous image $f(x, y)$ its derivative assumes a local maximum in the direction of the edge.

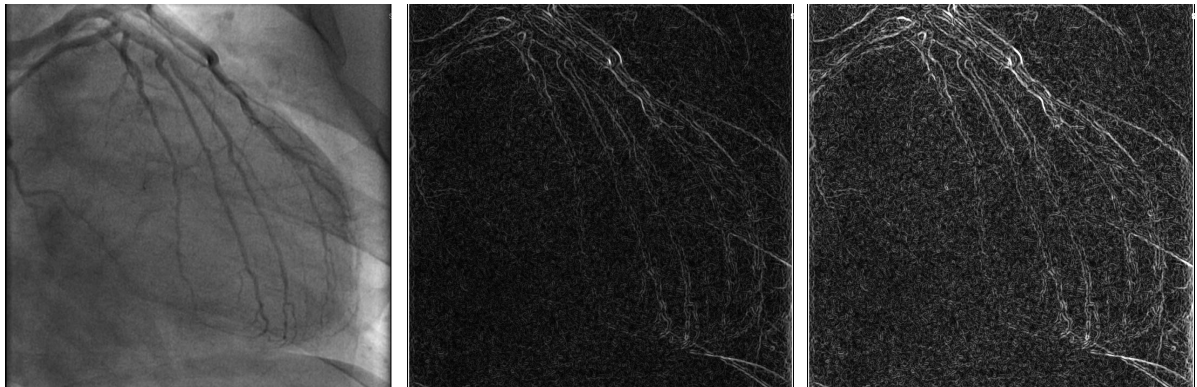

**Fig.4(a) Original CA Image (b)Edge detected Image using Eq(2.7) (c) Addition of image 4(a)&(b)**

Fig. 4 (b) is showing an output of edges of coronary arteries and it is produced by applying the Eq. (2.7) on the original image Fig. 4(a).
The image arithmetic is nothing but the addition of two images, that is Fig.4(a) and Fig.4(b) is added and the result is Fig. 4 (c).

$$f(x, y) = m(x, y) + n(x, y) \qquad (2.8)$$

Here $f(x, y)$ is an addition of each element in array m with the corresponding element in array n.

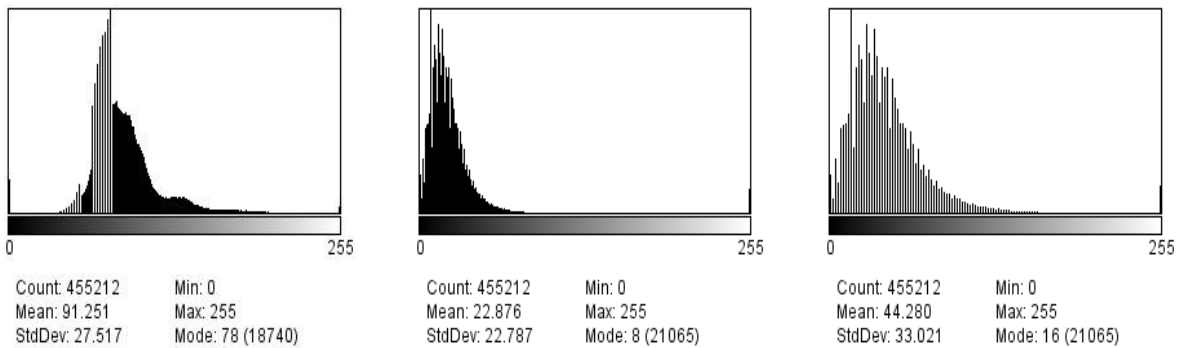

**Fig. 5(a) Histogram of Fig.4(a),    (b) Histogram of Fig. 4(b),    (c) Histogram of Fig. 4(c)**

The Fig. 5(a, b & c) are the histograms of Fig. 4(a, b and c) respectively. These histograms are clearly differentiating between the clearness and effective resolution of edges of coronary arteries.

**2.4 Image Shadowing**

The digital image measuring system (DIMS) which was developed for quantitative analysis on the surface damage, the shadow image processing technique (SIPT) was applied. By using both DIMS and SIPT, the surface damage on the contact could be successfully observed as the 3-D images[13].

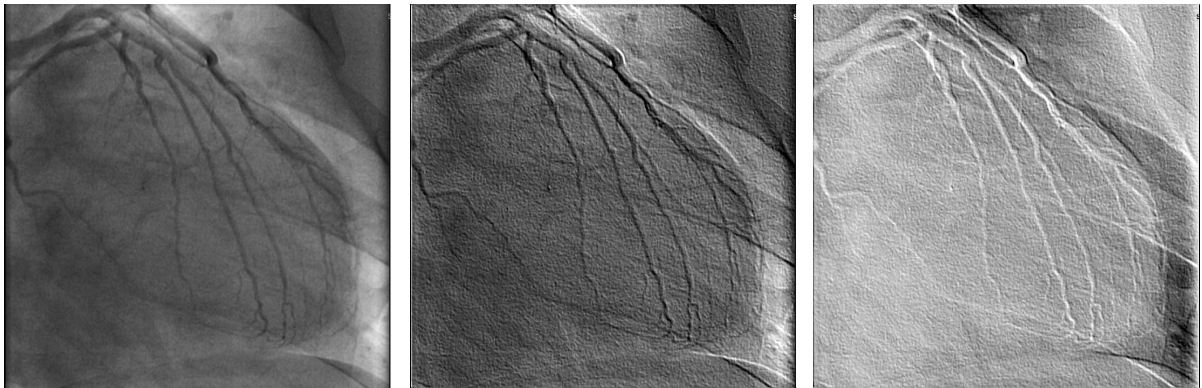

**Fig. 6(a) Original CA Image    (b) Image applied with shadow    (c) Shadow Invert image.**

Fig. 6 (b) is an output applied by shadowing affect on northeastern sides of detected edges of an original Coronary artery image. And the Fig. 6(b) clarifies the inverted sections of edges of coronary artery.

## 3. CONCLUSION

Coronary MRA is a rapidly evolving, new non-invasive technique. Although coronary MRA presently has limited clinical utility, it has the potential to aid in the diagnosis of coronary artery stenoses with a high degree of accuracy, especially in the proximal and middle segments, but remains challenging for distal coronary arteries. Magnetic resonance technology has some limitations, which makes it difficult to have better visualization of the coronary arteries. These limitations are secondary to the tortuous course of the coronary arteries, coronary arteries of smaller diameter (2.7 to 3.5 mm), rapid movement caused by respiratory and cardiac contractions, and the surrounding epicardial fat.[5,14] It may also be difficult to distinguish the coronary arteries from the parallel running coronary veins during the interpretation of coronary MRA, especially for the left circumflex coronary artery.

Future research should focus on the development of optimal respiratory compensation strategies by improving spatial and frequency domain filtering to visualize greater lengths of coronary arteries and faster acquisition of the data for better quality images.